\documentclass[runningheads]{llncs}
\usepackage{amssymb}
\usepackage{url}
\usepackage{hyperref}

\usepackage{graphicx}
\usepackage{amsmath}
\usepackage{scalerel}
\usepackage{tikz}
\usetikzlibrary{svg.path}
\definecolor{orcidlogocol}{HTML}{A6CE39}
\tikzset{
  orcidlogo/.pic={
    \fill[orcidlogocol] svg{M256,128c0,70.7-57.3,128-128,128C57.3,256,0,198.7,0,128C0,57.3,57.3,0,128,0C198.7,0,256,57.3,256,128z};
    \fill[white] svg{M86.3,186.2H70.9V79.1h15.4v48.4V186.2z}
                 svg{M108.9,79.1h41.6c39.6,0,57,28.3,57,53.6c0,27.5-21.5,53.6-56.8,53.6h-41.8V79.1z M124.3,172.4h24.5c34.9,0,42.9-26.5,42.9-39.7c0-21.5-13.7-39.7-43.7-39.7h-23.7V172.4z}
                 svg{M88.7,56.8c0,5.5-4.5,10.1-10.1,10.1c-5.6,0-10.1-4.6-10.1-10.1c0-5.6,4.5-10.1,10.1-10.1C84.2,46.7,88.7,51.3,88.7,56.8z};
  }
}
\newcommand\niceorcidID[1]{\href{https://orcid.org/#1}{\mbox{\scalerel*{
\begin{tikzpicture}[yscale=-1,transform shape]
\pic{orcidlogo};
\end{tikzpicture}
}{|}}}}
 
 \renewcommand\orcidID{\niceorcidID}

\usepackage{color}

\definecolor{liver}{RGB}{219,0,14} 
\definecolor{spleen}{RGB}{63,181,0} 
\definecolor{pancreas}{RGB}{11,100,255} 
\definecolor{gallbladder}{RGB}{255,204,0} 
\definecolor{unarybladder}{RGB}{0,196,255} 
\definecolor{rkidney}{RGB}{138,0,176} 
\definecolor{lkidney}{RGB}{255,43,78} 
\definecolor{rpsoas}{RGB}{255,185,93} 
\definecolor{lpsoas}{RGB}{102,205,170} 

\usepackage{pifont} 
\begin{document}
\title{Closing the Gap between Deep and Conventional Image Registration using Probabilistic Dense Displacement Networks}
\titlerunning{Probabilistic Dense Displacement Networks}
%
\author{Mattias P. Heinrich\inst{1}\orcidID{0000-0002-7489-1972}}
%
\authorrunning{M.P. Heinrich}
%
\institute{Institute of Medical Informatics, Universit\"{a}t zu L\"{u}beck
\email{heinrich@imi.uni-luebeck.de}, \url{https://github.com/multimodallearning/pdd_net}\\
}
\maketitle              
\begin{abstract}
Nonlinear image registration continues to be a fundamentally important tool in medical image analysis. Diagnostic tasks, image-guided surgery and radiotherapy as well as motion analysis all rely heavily on accurate intra-patient alignment. Furthermore, inter-patient registration enables atlas-based segmentation or landmark localisation and shape analysis. When labelled scans are scarce and anatomical differences large, conventional registration has often remained superior to deep learning methods that have so far mainly dealt with relatively small or low-complexity deformations. We address this shortcoming by leveraging ideas from probabilistic dense displacement optimisation that has excelled in many registration tasks with large deformations. We propose to design a network with approximate min-convolutions and mean field inference for differentiable displacement regularisation within a discrete weakly-supervised registration setting. By employing these meaningful and theoretically proven constraints, our learnable registration algorithm contains very few trainable weights (primarily for feature extraction) and is easier to train with few labelled scans. It is very fast in training and inference and achieves state-of-the-art accuracies for the challenging inter-patient registration of abdominal CT outperforming previous deep learning approaches by 15\% Dice overlap.


\end{abstract}
%
%
%

\section{Introduction and Related Work}
Conventional medical image registration mostly relies on iterative and multi-scale warping of a moving towards a fixed scan by minimising a dissimilarity metric together with a regularisation penalty. Deep learning based image registration (DLIR) aims to mimic this process by training a convolutional network that can predict the non-linear alignment function given two new unseen scans. Thus instead of multiple warping steps a single feed-forward transfer function has to be found using many convolution layers. The supervision for DLIR can be based on automatic or manual correspondences, semantic labels or intrinsic cost functions. It has immense potential for time-sensitive applications such as image-guidance, fusion, tracking and shape analysis through multi-atlas registration. However, due to the large space of potential deformations that can map two corresponding anatomies onto one another, the problem is much less constrained than image segmentation and therefore remains an open challenge. 

A number of approaches has been applied to brain registration \cite{balakrishnan2019voxelmorph,yang2017quicksilver}, which usually deals with localised deformations of few millimetres and for which huge labelled datasets ($\gg$100 scans) exist. For other anatomies in the abdomen, the prostate or lungs, with shape variations of several centimetres, DLIR was mainly applied to less complex cases of intra-patient registration \cite{hu2018weakly,krebs2018unsupervised}. For inhale-exhale lung registration the accuracy of DLIR is still inferior to conventional approaches: $\approx$ 2.5 mm in \cite{sentker2018gdl,de2019deep} compared to $<$1 mm in \cite{ruhaak2017estimation}. When training the state-of-the-art weakly-supervised DLIR approach Label-Reg \cite{hu2018weakly} on abdominal CT \cite{jimenez2016cloud} for inter-patient alignment, we reached an average Dice of only 42.7\%, which is still substantially worse than the conventional NiftyReg algorithm \cite{modat2010fast} with a Dice of 56.1\% and justifies further research.

Our hypothesis is that large and highly deformable transformations across different patients are difficult to model with a deep continuous regression network without resorting to complex multi-stage warping pipelines. Instead the use of discrete registration, which explores a large space of quantised displacements simultaneously, has been shown to capture abdominal and chest deformations more effectively \cite{heinrich2013towards,ruhaak2017estimation,xu2016evaluation} and can be realised with few or a single warping step. Unsurprisingly, discrete displacement settings have been explored in 2D vision for DLIR: namely the FlowNet-C \cite{dosovitskiy2015flownet}. A \textit{correlation layer} (see Eq. 1 in \cite{dosovitskiy2015flownet}) is proposed that contains no trainable weights and computes a similarity metric of features from two images by shifting the moving image with a densely quantised displacement space ($21\times21$ pixel offsets) yielding a 441-channel joint feature map. Next, a very large  $441(+32)\times256\times3\times3$ kernel is learned (followed by further convolutions) that disregards the explicit 4D geometry of the displacement space. Hence, the large number of optimisable parameters results in huge requirements of supervised training data. Extending this idea to 3D is very difficult as the dimensionality increases to 6D after dense correlation and has not been yet considered despite its benefits. Probabilistic and uncertainty modelling has been studied in DLIR, cf. \cite{krebs2018unsupervised,yang2017quicksilver}, but not in a discrete setting.

\subsubsection*{Contributions}
We propose a new learning model for DLIR that better leverages the advantages of probabilistic dense displacement sampling by introducing strong regularisation with differentiable constraints that explicitly considers the 6D nature of the problem. We hence decouple convolutional feature learning from the fitting of a spatial transformation using mean-field inference for regularisation \cite{krahenbuhl2011efficient,zheng2015conditional} and approximate min-convolutions \cite{felzenszwalb2006efficient} for computing inter-label compatibilities. Our feature extractor uses 3D deformable convolutions \cite{heinrich2019obelisk} and is very lightweight. To our knowledge this is the first approach that combines discrete DLIR with the differentiable use of mean-field regularisation. In contrast to previous work, our model requires fewer trainable weights, captures larger deformations and can be trained from few labelled scans to high accuracy. We also introduce a new non-local label loss for improved guidance instead of the more widely used spatial transformer based loss.
  

\begin{figure}[t]
\centering
\includegraphics[width=\linewidth]{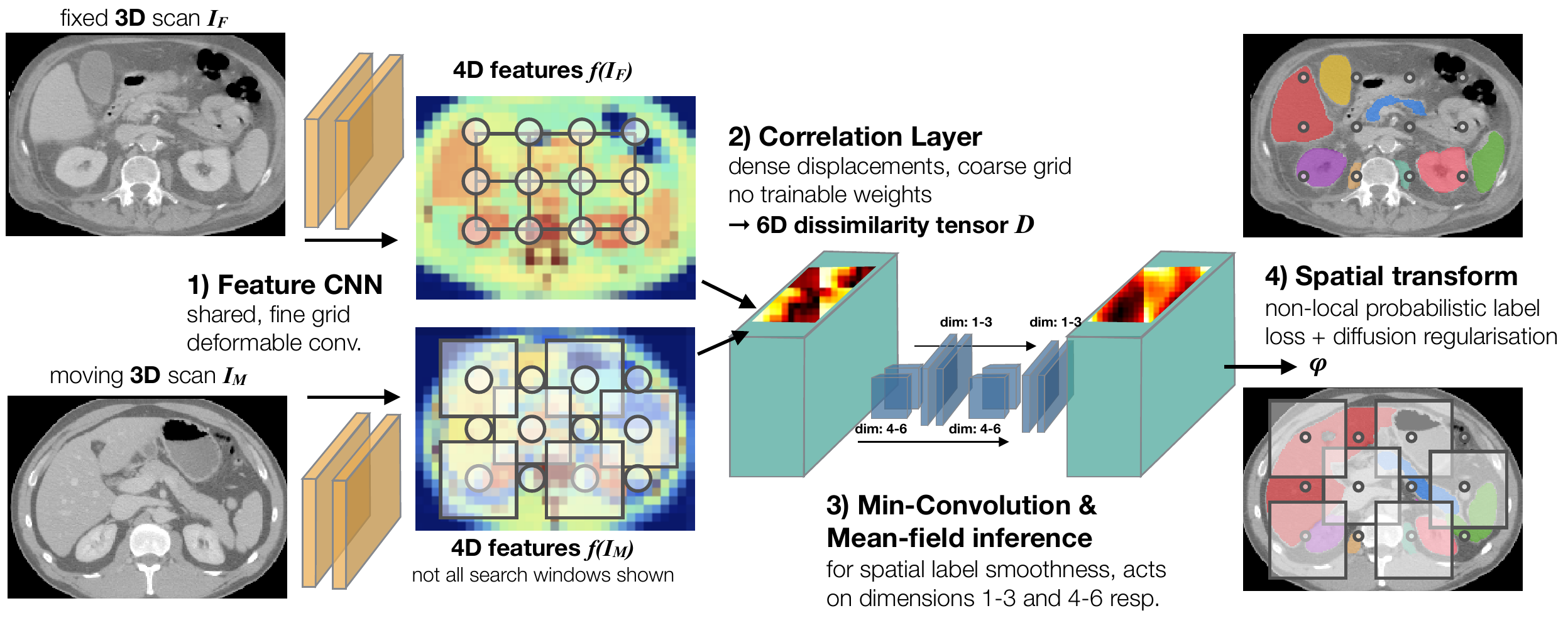}
\caption{Concept of probabilistic dense displacement network: 1) deformable convolution layers extract features for both fixed and moving image. 2) the correlation layer evaluates for each 3D grid point a dense displacement space yielding a 6D dissimilarity map. 3) spatial filters that promote smoothness act on dimensions 4-6 (min-convolutions) and dim. 1-3 (mean-field inference) in alternation. 4) the probabilistic transform distribution obtained using a softmax (over dim. 4-6) is used in a non-local label loss and converted to 3D displacements for a diffusion regularisation and to warp images.}
\label{fig:overview}
\end{figure}

\section{Methods}\label{sec:methods}
We aim to align a fixed $I_F$ and moving $I_M$ 3D scan by finding a spatial transformation $\varphi$ based on a learned feature mapping $f$ of $I_F$ and $I_M$ subject to constraints on the regularity of $\varphi$. In order to learn a suitable feature extraction that is invariant to noise and uninformative contrast-variations, we provide a supervisory label during training for both volumes $\mathbf{\ell}_F$ and $\mathbf{\ell}_M$, for which $\mathbf{\ell}_F\approx \varphi\circ \mathbf{\ell}_M$ should hold after registration. We define spatial coordinates as continuous variables $\mathbf{x}\in(-1,+1)^3$ and use trilinear interpolation to sample from discrete grids. $\varphi$ is parameterised with a set of $\mathbf{k}\in|K|\in\mathbb{R}^3$ (a few thousands) control points on a coarser grid. The range of displacements $\mathbf{d}$ is constrained to a discrete displacement space, with linear spacing e.g. $\mathcal{L}=q\cdot\{-1,-\frac{6}{7},-\frac{5}{7},\ldots,+\frac{5}{7},+\frac{6}{7},+1\}^3$, where $q$ is a scalar that defines the capture range and in our case $|\mathcal{L}|$ is $3375$. The network model should predict a 6D tensor of displacement probabilities $K \in\mathbb{R}^3\times\mathcal{L}\in\mathbb{R}^3$, where the sum over the dimensions 4-6 of $\mathcal{L}$ for each control point is 1. The (inner) product of the probabilities with the displacements $\mathcal{L}$ yields the weighted average of these probabilistic estimates to obtain 3D displacements for $\varphi$ during inference. 

\textbf{1) Convolutional feature learning network: } To learn a meaningful nonlinear mapping from input intensities to a dense feature volume (with $|c|=16$ channels and a stride of 3), we employ the Obelisk approach \cite{heinrich2019obelisk}, which comprises a 3D deformable convolution with trainable offsets followed by a simple $1\times1$ MLP and captures spatial context very effectively. We extend the authors' implementation by adding a normal $5\times5\times5$ convolution kernel with 4 channels prior to the Obelisk layer to also learn edge-like features. The network has 64 spatial filter offsets and in total 120k trainable parameters, shared for fixed and moving scan to yield $f(I_F)$ and $f(I_M)$.

\textbf{2) Correlation layer for dense displacement dissimilarity: }Given the feature representation of the first part, we aim to find a regularised displacement field that assigns a vector $\mathbf{d}$ to every control point for a nonlinear transform $\varphi(\mathbf{k})\leftarrow\mathbf{d}$ that maximises the (label) similarity between fixed and warped moving scan. As done in conventional discrete registration \cite{heinrich2013towards} and the correlation layer of \cite{dosovitskiy2015flownet}, we perform a dense evaluation of a similarity metric over the displacement search space $\mathbf{d}\in\mathcal{L}$. The negated mean squared error (MSE) across the feature dimension $c$ of learned descriptors is used to obtain the 6D tensor of dissimilarities $\mathcal{D}(\mathbf{k},\mathbf{d})=-\frac{1}{|c|}\sum_{c}(f_c(I_F)_{\mathbf{k}}-f_c(I_M)_{\mathbf{k+d}})^2$. Different metrics such as the correlation coefficient could be employed. Due to the sparsity of the control points the displacement similarity evaluation requires less than 2 GFlops in our experiments. The capture range of displacements $q$ is set to 0.4.

\textbf{3) Regularisation using min-convolutions and mean-field inference:} Since nonlinear registration is usually ill-posed, additional priors are used to keep deformations spatially smooth. In contrast to other work on DLIR, which in principle learn an unconstrained deformation and only enforce spatial smoothness as loss term, we propose to model regularisation constraints as part of the network architecture. A diffusion-like regularisation penalty for displacements based on their squared difference $\mathcal{R}(\mathbf{d}_i,\mathbf{d}_j)=||\mathbf{d}_i-\mathbf{d}_j||^2$ is often used in Markov random field (MRF) registration \cite{felzenszwalb2006efficient} and e.g. optimised with loopy belief propagation (LBP). \cite{kamnitsas2017efficient} and  \cite{zheng2015conditional} integrated smoothness constraints of graphical models into end-to-end learned segmentation networks. Since, LBP requires many iterations to yield an optimum and is hence not well suited as unrolled network layers, we use the fast mean-field inference (two iterations) used for discrete optimisation in \cite{krahenbuhl2011efficient} (in \cite{zheng2015conditional} 5 iterations were used). It consists of two alternating steps: a label-compatibility transform that acts on spatial control points independently and a filter-based message passing implemented using average pooling layers with a stride of 1. 

As noted in \cite{felzenszwalb2006efficient} the diffusion regularisation for a dense displacement space can be computed using min-convolutions with a lower envelope of parabolas rooted at the (3D) displacement offsets with heights equalling to the sum of dissimilarity term and the previous iteration of the mean-field inference. This lower envelope is not directly differentiable, but we can obtain a very accurate approximation using first, a min-pooling (with stride=1) that finds local minima in the cost tensor followed by two average pooling operations (with stride=1) that provide a quadratic smoothing. As shown with blue blocks in Fig.~\ref{fig:overview}, the novel regularisation part of our approach comprises min- and average-pooling layers that act on the 3 displacement dimensions (min-convolution) followed by average filtering on the 3 spatial dimensions (mean-field inference). Before each operation, scaling and bias factors $\alpha_1-\alpha_6$ are introduced and optimised together with the feature layers during end-to-end training. 

\textbf{Probabilistic transform losses and label supervision: }We can make further use of the probabilistic nature of our displacement sampling and specify our supervised label loss term based on a non-local means weighting \cite{rousseau2011supervised}. I.e., we first convert the negated output of the regularisation part (scaled by $\alpha_6$) into pseudo-probabilities using a softmax computed over the displacements. Next, one-hot representations of the moving segmentation are sampled at the same spatially displaced locations and these vectors are multiplied by the estimated probabilities to compute the label loss as MSE w.r.t. the ground truth (one-hot) segmentation. The continuous valued 3D displacement field $\varphi$ is obtained by a weighted average of the probabilistic estimates multiplied with the displacement labels and followed by trilinear interpolation to the image resolution. A diffusion regularisation penalty over all 3 spatial gradients $\lambda\cdot(|\nabla\varphi_1|^2+|\nabla\varphi_2|^2+|\nabla\varphi_3|^3)$ of the displacement field is employed to enable a user-defined balancing between a smooth transform (with low standard deviation of Jacobians) and accurate structural alignment.



\section{Experimental Validation}\label{sec:experiments}

To demonstrate the ability of our method to capture very large deformations across different patients within the human abdomen, we performed experiments with a 3-fold cross validation on 10 contrast-enhanced 3D CT scans of the VISCERAL3 training data \cite{jimenez2016cloud} with each nine anatomical structures manually segmented: \textcolor{liver}{$\blacksquare$} liver, \textcolor{spleen}{$\blacksquare$} spleen, \textcolor{pancreas}{$\blacksquare$} pancreas, \textcolor{gallbladder}{$\blacksquare$} gallbladder, \textcolor{unarybladder}{$\blacksquare$} unary bladder,  \textcolor{rkidney}{$\blacksquare$} right kidney, \textcolor{lkidney}{$\blacksquare$} left kidney, \textcolor{rpsoas}{$\blacksquare$} right psoas major muscle (psoas) and \textcolor{lpsoas}{$\blacksquare$} left psoas (see Fig. \ref{fig:result}). The images were resampled to isotropic voxel sizes of 1.5 mm$^3$ with dimensions of $233\times168\times286$ voxels and without any manual pre-alignment. 

We compare our \textbf{p}robabilistic \textbf{d}ense \textbf{d}isplacement network (\textbf{pdd-net})\footnote{our code with all implementation details will be made publicly available.} with the two conventional algorithms \textbf{NiftyReg} \cite{modat2010fast} and \textbf{deeds} \cite{heinrich2013towards} that performed best in the inter-patient abdominal CT registration study of \cite{xu2016evaluation}, a task not yet tackled by DLIR. NiftyReg was used with mutual information and a 5-level multi-resolution scheme to capture large deformations and has a run-time of 40-50 sec. Deeds was considered with a single scale dense displacement space (which takes about 4-6 sec) and then extended to three-levels of discrete optimisation (25-35 sec run-time). Next, we trained the weakly-supervised DLIR method \textbf{Label-Reg} \cite{hu2018weakly} on our data (in $>$24 hours per fold). To reduce memory requirements below 32 GBytes, the resolution was reduced to 2.2 mm and the base channel number halved to 16. Further small adjustments were made to optimise for inter-patient training. We implemented a 3D extension of \textbf{FlowNet-C} \cite{dosovitskiy2015flownet} in pytorch with Obelisk feature extraction, a dense correlation layer and a regularisation network that has $|\mathcal{L}|=3375$ input channels, comprises five 3D conv. layers with batch-norm and PReLU. It has  2 million weights and outputs a (non-probabilistic) 3D displacement field. In order to obtain meaningful results it was necessary to add a semantic segmentation loss to the intermediate output of the Obelisk layers. Our proposed method employs the same feature learning part (with 200k parameters) but now uses min-convolutions, mean-field inference (no semantic guidance) and the non-local label loss, which adds only 6 trainable weights (and not 2 million). The influence of these three choices is analysed with an ablation study, where a replacement of Obelisk feature learning with handcraft self-similarity context features \cite{heinrich2013towards} is also considered. We use a diffusion regularisation weight of $\lambda=1.5$ for control grids of size $32^3$ and affine augmentation of fixed scans throughout and trained our networks with Adam (learning rate of 0.01) for 1500 iterations in $\approx$90 minutes and $\approx$16 GByte of GPU memory with checkpointing. We implemented an instance-wise gradient descent optimiser that refines the feed-forward predictions. \cite{balakrishnan2019voxelmorph} also used this idea, but in our case it is a hundred times faster (0.24 sec. vs 24 sec.), since we can directly operate on the pre-computed displacement probabilities and require no iterative back-propagation through the network.  
\begin{table}[tb]
\small
  \caption{Quantitative comparative evaluation of cross-validation for 10 scans of the VISCERAL anatomy 3 dataset, based on 24 combinations of test scans not seen in training (numbers are Dice scores). Our method \textbf{pdd-net} outperforms the considered DLIR methods, Label-Reg and FlowNet-C, by a margin of about 15\% and closes the gap to conventional methods, NiftyReg and deeds, for this task. Our ablation study shows the benefits of 1) the use of learned Obelisk features vs handcrafted self-similarity context (SSC) descriptors, 2) employing mean-field inference and 3) the use of our new non-local label loss. $^\circ$Additionally a fast instance level optimisation was implemented for \textbf{pdd-net+inst.}. $^*$FlowNet-C is our 3D extension of \cite{dosovitskiy2015flownet} with Obelisk features and a trainable regularisation network. To compare: the Dice before registration was 30.0\% on average.}
  \label{tab:results}
  \centering
  \resizebox{\textwidth}{!}{\begin{tabular}{l|ccc|ccccccccc|lcc}
  \textbf{Method}  &1)&2)&3)& \textcolor{liver}{$\blacksquare$} & \textcolor{spleen}{$\blacksquare$} & \textcolor{pancreas}{$\blacksquare$} & \textcolor{gallbladder}{$\blacksquare$} & \textcolor{unarybladder}{$\blacksquare$} & \textcolor{rkidney}{$\blacksquare$} & \textcolor{lkidney}{$\blacksquare$} & \textcolor{rpsoas}{$\blacksquare$} & \textcolor{lpsoas}{$\blacksquare$} & average & std(Jac) & runtime\\
   \hline
Label-Reg &&&&71&51&7&5&38&53&59&45&55&42.7$\pm$5.5&0.58&4 sec.\\

\hline
FlowNet-C$^*$ &\ding{52}&$$&\ding{52}&73&45&9&7&40&48&53&49&52&41.8$\pm$7.2&0.34&0.13 sec.\\
pdd+SSC &\ding{56}&\ding{52}&\ding{52}&69&47&8&6&49&56&56&59&57&45.2$\pm$5.4&0.54&1.38 sec.\\
pdd w/o MF &\ding{52}&\ding{56}&\ding{52}&74&53&7&8&49&65&63&56&60&48.2$\pm$4.8&0.38&0.45 sec.\\
pdd w/o NL &\ding{52}&\ding{52}&\ding{56}&83&62&11&8&47&69&68&60&60&51.9$\pm$7.1&0.39&0.57 sec.\\
\textbf{pdd-net} &\ding{52}&\ding{52}&\ding{52}&84&62&13&12&57&76&70&69&68&\textbf{56.7$\pm$6.0}&0.40&0.57 sec.\\
\textbf{pdd+inst.$^\circ$} &\ding{52}&\ding{52}&\ding{52}&83&66&18&14&58&77&74&68&68&\textbf{58.4$\pm$5.9}&0.26&0.71 sec.\\

\hline
deeds+SSC &\multicolumn{3}{c|}{1 level}&72&50&14&13&51&54&58&62&60&48.0$\pm$6.8&0.67&4 sec.\\
deeds+SSC &\multicolumn{3}{c|}{3 level}&78&62&18&19&60&71&67&70&69&57.0$\pm$8.2&0.27&25 sec.\\
NiftyReg NMI &\multicolumn{3}{c|}{5 level}&77&58&19&27&56&70&65&67&66&56.1$\pm$18&1.30&42 sec.\\
\hline
  \end{tabular}}
\end{table}


\section{Results and Discussions}\label{sec:discussion}
The inference time of \textbf{pdd-net} is only 0.57 sec, yielding plausible displacement fields with a standard deviation of the Jacobian determinants of 0.40 and $< $1\% folding voxels (negative Jacobians). Table~\ref{tab:results} shows the average Dice scores across 24 registrations of the cross-validation, where no labelled training scan was used for any evaluated test registration. Our method outperforms the two compared DL approaches,\textbf{ Label-Reg} and \textbf{FlowNet-C}, by a margin of about 15\% points and achieves 56.7\% Dice for this challenging inter-patient task with an initial alignment of only 30\%. It is  10\% better than a comparable setting of the conventional discrete registration \textbf{deeds} with one grid-level. In particular the labels \textcolor{liver}{$\blacksquare$}, \textcolor{spleen}{$\blacksquare$}, \textcolor{unarybladder}{$\blacksquare$}, \textcolor{rkidney}{$\blacksquare$}, \textcolor{lkidney}{$\blacksquare$}, \textcolor{rpsoas}{$\blacksquare$} and \textcolor{lpsoas}{$\blacksquare$} are very well aligned. Our instance-wise (per scan-pair) optimisation requires 0.24 sec, reduces foldings (to less than 0.6\%) and further increases the accuracy to 58.4\%, which is above the level of the conventional multi-level registrations \textbf{deeds} and \textbf{NiftyReg}.

Comparing deeds+SSC with one grid-level to our variant pdd+SSC, which uses the same self-similarity features and only adapts the $\alpha$ parameters of the regularisation part, we get a similar accuracy and deformation complexity. This suggests that the proposed regularisation layers with min-convolutions and two mean-field inference steps can nearly match the capabilities of the full sequential MRF optimisation in \cite{heinrich2013towards}. Using weak supervision to learn features results in more than 20\% increased Dice. The non-local loss term and our instance-wise fine-tuning, contribute further gains of 5\% and 2\% Dice overlap, respectively. The importance of the mean-field inference is clear, given the inferior quality of an unconstrained FlowNet-C with more trainable weights or our variant that only uses min-convolutions but no filtering in spatial domain. We achieve a more robust alignment quality (lower std.dev. of Dice) than conventional registration. Visual registration examples are shown in Fig.~\ref{fig:result} and as surface-rendered video files in the supplementary material.


\begin{figure}[bt]
\centering
\includegraphics[width=\linewidth]{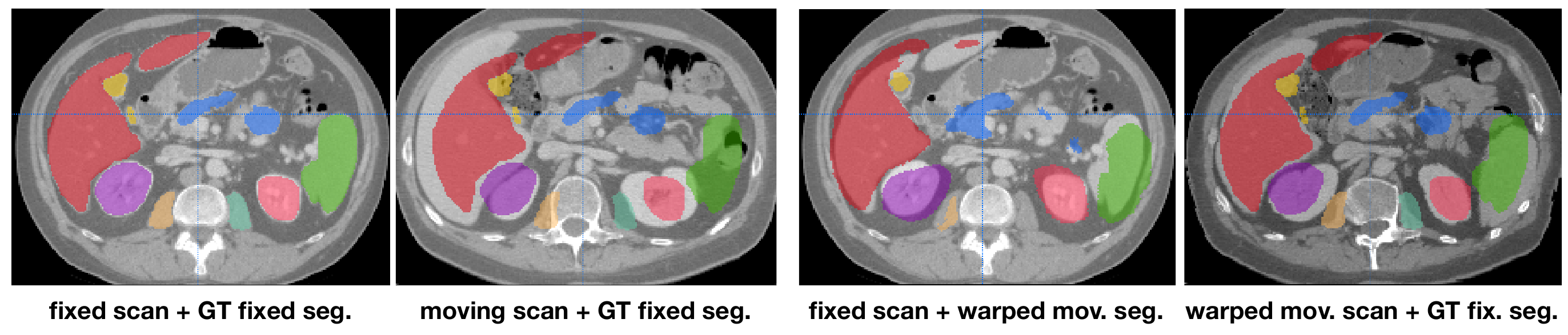}
\caption{Visual outcome of proposed \textbf{pdd-net} method to register two patients and transfer a segmentation (moderate example). Most organs have been very well aligned and also anatomies that are not labelled in training (stomach, vertebras) can be registered.}
\label{fig:result}
\end{figure}

\section{Conclusion}
Our novel \textbf{pdd-net} combines probabilistic dense displacements with differentiable mean-field regularisation to achieve one-to-one accuracies of over 70\% Dice for 7 larger anatomies for inter-patient abdominal CT registration. It outperforms the previous deep learning-based image registration (DLIR) methods, Label-Reg and FlowNet-C, by a margin of 15\% points and can be robustly trained with few labelled scans. It closes the quality gap of DLIR (with small training datasets) to state-of-the-art conventional methods, exemplified by NiftyReg and deeds, while being extremely fast (0.5 sec). Our concept offers a clear new potential to enable the use of DLIR in image-guided interventions, diagnostics and atlas-based shape analysis beyond the currently used pixel segmentation networks that lack geometric interpretability. Future work could yield further gains by using multiple alignment stages and a more adaptive sampling of control points. A more elaborate evaluation on larger datasets with additional evaluation metrics (surface distances) could provide more insights into the method's strengths and weaknesses.   

  %
%
%
\bibliographystyle{splncs04}
\bibliography{miccai2019_pdd.bib}
\end{document}